\documentclass[letterpaper]{article} 
\usepackage{aaai25}  
\usepackage{times}  
\usepackage{helvet}  
\usepackage{courier}  
\usepackage[hyphens]{url}  
\usepackage{graphicx} 
\usepackage{natbib}  
\usepackage{caption} 
\graphicspath{{./}{./plots/}}
\usepackage{algorithm}
\usepackage{algorithmic}
\usepackage{newfloat}
\usepackage{listings}
\DeclareCaptionStyle{ruled}{labelfont=normalfont,labelsep=colon,strut=off} 
\floatstyle{ruled}
\newfloat{listing}{tb}{lst}{}
\floatname{listing}{Listing}
\title{Reorganizing attention-space geometry with expressive attention}

\author{Claudius Gros}
\affiliations{
Institute for Theoretical Physics \\
Goethe University Frankfurt\\
Frankfurt a.~M., Germany \\
\texttt{\{gros07\}@itp.uni-frankfurt.de} \\
}

\begin{document}

\maketitle

\begin{abstract}
Attention involves comparing query and key 
vectors in terms of a scalar product, 
$\mathbf{Q}^T\mathbf{K}$, together with a 
subsequent softmax normalization. Classically,
parallel/antiparallel queries and 
keys lead to large/small attention weights.
Here we study expressive attention (EA), 
which is based on $(\mathbf{Q}^T\mathbf{K})^2$, 
the squared dot product. In this case attention 
is enhanced when query and key are either parallel 
or antiparallel, and suppressed for orthogonal 
configurations. EA can be introduced into any 
attention-based code without additional
compute costs or memory requirements. 
For a series of autoregressive prediction tasks, 
we find that EA performs at least 
as well as the standard mechanism, dot-product 
attention (DPA). Increasing task complexity, 
EA is observed to outperform DPA with increasing 
margins, which also holds for multi-task settings.
For a given model size, EA manages to achieve 
100\% performance for a range of complexity 
levels not accessible to DPA.
\end{abstract}

\section{Introduction}

\begin{figure}[t]
\centerline{
\includegraphics[width=0.48\linewidth]{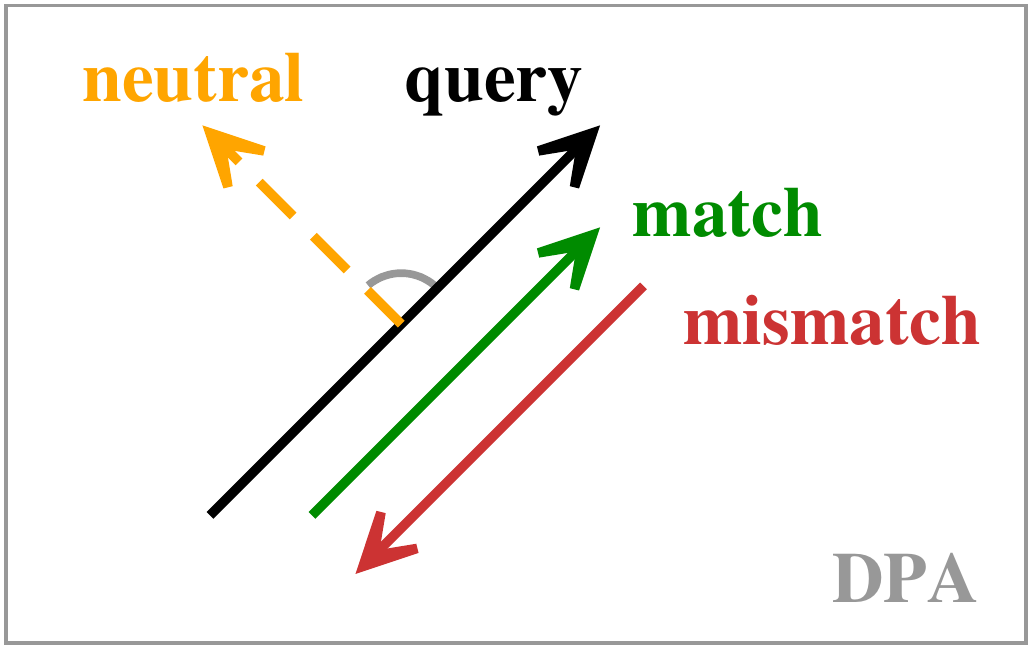}
\hfill
\includegraphics[width=0.48\linewidth]{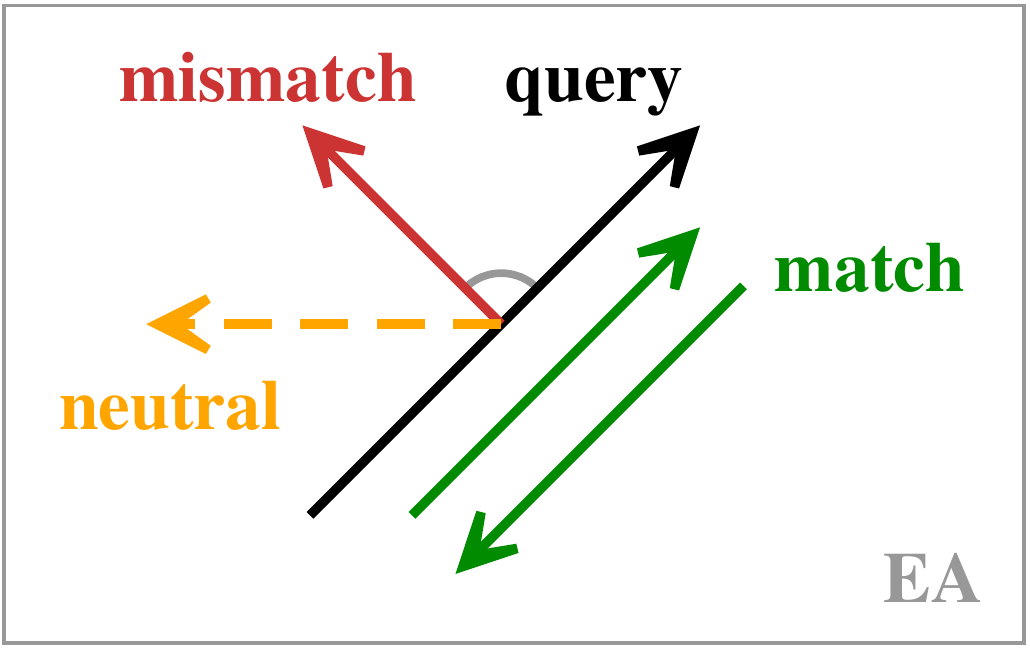}
}
\caption{Comparing keys (colored arrows) with
a given query (black arrow), constitutes the basis 
of ML attention. Attention weights are large/small
when query and key match/mismatch (green/red), or 
in between, viz neutral (dashed yellow).
For the standard dot-product attention 
(DPA, left panel, see (\ref{DPA_def})), attention weights are large/small
for parallel/antiparallel keys and query. For expressive
attention (EA, right panel, defined in (\ref{REA_def})),
both parallel and antiparallel alignments form matching 
pairs, with orthogonal keys and queries leading to a mismatch.
}
\label{fig:attention_illustration}
\end{figure}

Since its inception \citep{bahdanau2014neural,vaswani2017attention},
the attention mechanism has had a transformative
impact on machine learning \citep{soydaner2022attention,minaee2024large}.
At its core, attention facilitates pairwise information routing 
between tokens, with information being transmitted when a 
given matching condition is fulfilled. For the latter
query and key vectors $\mathbf{Q}$ and $ \mathbf{K}$
are compared in terms of the respective dot products,
$\mathbf{Q}^T\mathbf{K}$, which are mapped to a
normalized probability distribution via a
subsequent softmax operation. As illustrated in
Fig.~\ref{fig:attention_illustration}, this
setup, dot-product attention (DPA), constrains
the central part of the matching condition to
a one-dimensional subspace of the otherwise
potentially large space of attention heads. Given
that attention matrices tend to be sparse
\citep{likhosherstov2023expressive}, it 
might be favorable to allow the system
to express low attention states
in a larger subspace.

Here we introduce and discuss a modified 
attention mechanism, expressive attention 
(EA), which is based on the presumption
that low attention should correspond to orthogonal 
query and key configurations. This venue
allows attention to express itself in the
entire attention space, as illustrated in
Fig.~\ref{fig:attention_illustration}.
All one needs to realize EA is to base
attention on $(\mathbf{Q}^T\mathbf{K})^2$,
the squared scalar product. In practice,
this corresponds to changing a single line 
of code, which makes it straightforward to 
introduce effective attention into production
codes. Computational complexity and memory
requirements are not affected.

We present a comparative evaluation of classical
DPA and EA. For this purpose we employ a suite
of autoregressive prediction tasks (NT prediction 
tasks) that are based on time series generated 
by the delayed addition of two or more numbers,
where addition is modulo a generic basis $N$, like 
2 or 16 (binary/hexadecimal case). NT tasks are
generalized delayed XOR tasks that become 
increasingly difficult to solve for larger $N$ 
and delays $\tau$ \citep{schubert2021local}.
Models are expected to achieve not just a good 
performance, but 100\%, which reflects the
view that systems can be said to exhibit 
reasoning errors when achieving less than 100\% 
accuracy for solvable tasks 
\citep{liu2024exposing}. The question is
then, how long perfect performance can be maintained 
when systematically increasing task difficulty,
viz complexity.

\section{Related work}

Alternatives to the standard, attention-based
transformer have been studied intensively 
during the past years. At a basic level, 
non-transformer based architectures for 
sequence processing have been proposed, like
structured and/or selective state sequence models 
\citep{gu2021efficiently,gu2023mamba}. Other
approaches aim to reduce compute costs, for example
by making use of a  multi-to-multi recurrent
network analogy \citep{feng2024attention}. Another
route is linearized attention 
\citep{katharopoulos2020transformers,wang2020linformer,wu2022flowformer}
for which compute scales linearly with context length,
and not quadratically. In practice, linear attention 
comes with various problems of its own, like 
unbounded gradients and attention dilution
\citep{qin2022devil,han2023flatten}. 
In general, linearized attention models substitute
the softmax operation of classical dot-product
attention by various linear kernel functions, 
which may have stochastic components 
\citep{choromanski2020rethinking,peng2021random},
well designed spatial dependencies \citep{qin2022cosformer},
or being derived from a singular value decomposition
\citep{chen2024primal}. 

A different aspect is addressed
by rotary positional embedding (RoPE), 
which adds positional information directly to the 
attention mechanism, via position-dependent rotations
of query and key vectors
\citep{su2021roformer,su2024roformer}. RoPE
can be used in particular for extending
context windows \citep{chen2023extending}.
Somewhat similar in spirit is $(IA)^3$, which
uses element-wise rescaled key and values vectors 
(in addition to rescaled hidden layer activities) 
for fine-tuning pretrained LLMs for downstream 
tasks \citep{liu2022few}.

Two main routes are available for testing the
performance of sequence processing architectures, 
such as transformers. The first is to use databases 
relevant for real-world applications
\citep{kaddour2023challenges,yang2024harnessing},
the second is to rely on synthetic test suites.
The latter approach is used standardly when
studying learning biases, e.g., when comparing
length generalization scores for functions 
like `parity', `majority', `first' or `mean'
\citep{abbe2023generalization,hahn2024sensitive}.
Synthetic test environments have been employed also 
for the study of attention glitches in reasoning 
tasks \citep{liu2024exposing}. 
For our studies we use NT tasks, a suite of 
synthetic autoregressive prediction tasks 
which can be tuned to a desired level of 
difficulty, viz complexity. We will study
in particular the transition between good and
optimal performance when increasing context 
length.

\section{Background}

Multiplying token activity with the respective
query, key and value matrices generates
three vectors, $\mathbf{Q}_i$,
$\mathbf{K}_i$ and $\mathbf{V}_i$, 
specifically for each token $i$ in
a given attention layer. The activity 
$\mathbf{y}_m$ of token $m$ is given by
$\mathbf{y}_m=\sum_{k\le m} a_{mk}\mathbf{V}_k$,
when masked attention is used. The 
attention matrix $a_{mk}\ge0$ is normalized
row-wise, $1=\sum_k a_{mk}$, encoding 
how much information is transferred from 
token $k$ to token $m$. This setup implements 
information routing for any suitable 
mechanism determining the individual
$a_{mk}$. The standard approach 
\citep{vaswani2017attention},
\begin{equation}
a_{mk}\big|_{\rm DPA} = \frac{1}{Z_m}
\exp\left(\beta\mathbf{Q}_m^T\mathbf{K}_k\right)\,
\label{DPA_def}
\end{equation}
takes the scalar product $\mathbf{Q}^T\mathbf{K}$
as the fundamental similarity measure. The softmax
operation included in (\ref{DPA_def}) transforms 
the basic similarity measure into a probability 
distribution function, with $Z_m$ being an appropriate 
normalization factor. One can set $\beta=1$ 
without affecting performance, as done here, 
as the standard expression $\beta=1/\sqrt{N_{\rm con}}$ 
only affects the scaling of weight and query matrices 
with context length, 
$N_{\rm con}$ \citep{vaswani2017attention}. 

As an alternative to (\ref{DPA_def}) we investigate
expressive attention (EA), defined by
\begin{equation}
a_{mk}\big|_{\rm EA} = \frac{1}{N_m}
\frac{z_{mk}^2}
     {1+z_{mk}^2}, \quad\quad
z_{mk} = \mathbf{Q}_m^T\mathbf{K}_k\,,
\label{REA_def}
\end{equation}
where $N_m$ is a normalization factor.
An illustration is presented in
Fig.~\ref{fig:attention_illustration}.
Importantly, EA is a function of the
squared dot-product $z_{mk}$. The difference
between EA and DPA can be expressed in
terms of attention space geometry.

\paragraph{Attention space geometry}
Attention mechanisms based on comparing keys
and query vectors allocate semantic meanings 
to the respective geometric configurations.
Classically (\ref{DPA_def}), parallel/anti-parallel
keys and queries lead to high/low attention weights.
For a $d$-dimensional attention-head space, the
$d-1$ possible orthogonal arrangements lead
to intermediate attention weights.

Full attention matrices tend to be sparse
\citep{likhosherstov2023expressive},
which allows in many instances to reduce
the dimensionality of attention matrices
from the start via a dedicated decimation process
\citep{roy2021efficient,lou2024sparser}. This
observation suggest that it may be advantageous
if all possible $d-1$ dimensional orthogonal
configurations would lead to reduced attention 
weights. Eq.~(\ref{REA_def}) encodes this concept,
as illustrated in Fig.~\ref{fig:attention_illustration}.
The naming of our approach, `expressive attention',
is motivated by above geometric considerations.

\begin{figure}[t]
\centerline{
\includegraphics[width=0.70\linewidth]{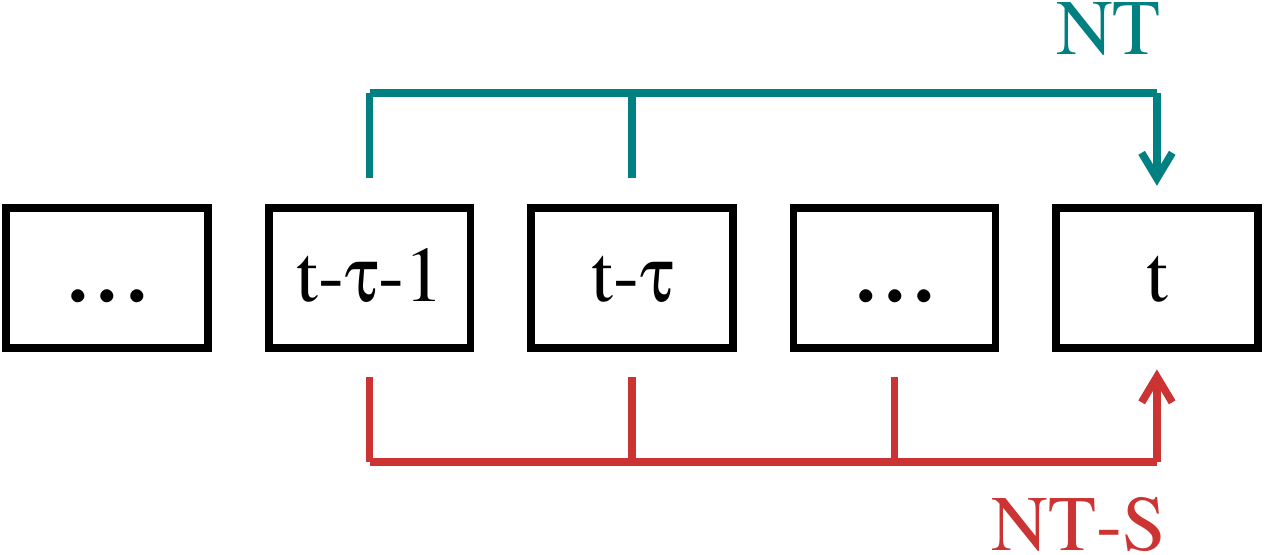}
}
\caption{For a time series of symbols (boxes),
the structure of the NT prediction task (teal), 
as defined by (\ref{NT_time}), and of the NT-S
variant (red), see (\ref{NT-S_time}). 
}
\label{fig:NT_illustration}
\end{figure}

\section{Experimental settings}

\paragraph{Suite of Tasks}
A basic non-linear autoregressive task is to 
predict the next token of a time series
\citep{li2024mechanics}. For our investigation
we are interested in a suite of time series
prediction tasks which allows to systematically
increase and tune difficulty.
As a motivation we first consider
the basic XOR setting \citep{gros24complex}.
\begin{eqnarray}
\nonumber
x(t) &=& \mathrm{XOR}\big(x(t-1),x(t-2)\big) \\
&=& \big[x(t-1)+x(t-2)\big]\%2\,,
\label{XOR_time}
\end{eqnarray}
where we used in the second step that the XOR
operation correspond to the addition of two boolean
inputs $x=0/1$, modulo two. As illustrated in
Fig.~\ref{fig:NT_illustration}, we generalize 
(\ref{XOR_time}) to the case of a general delay
$\tau\in[1,2,3,..]$ and basis $N\in[2,3,4,..]$,
\begin{equation}
\hbox{\sc NT:}\quad
x(t) = \big[x(t-\tau)+x(t-1-\tau)\big]\%N\,,
\label{NT_time}
\end{equation}
which defines the NT prediction task.
For the XOR series, recovered for N2T1, two
types of cyclic patterns are generated,
\begin{equation}
011011011011\dots,
\qquad\quad
000000000000\dots\,.
\label{XOR_cycles}
\end{equation}
There are four initial conditions, 11, 01, 10 and 00,
of which the first three give rise to cycles of 
type 011, with the last leading to the default series.
The complexity of the associated prediction task
increases systematically when increasing N and/or 
$\tau$. For example, one has
\begin{equation}
512\cdot\underline{120} + 
 64\cdot\underline{60} + 
  8\cdot\underline{30} + 
  1\cdot\underline{15} + 
  1\cdot\underline{1}
=  65536 = 16^4
\label{N16T3_all}
\end{equation}
for N16T3, which states that there are
512/64/8/1/1 cycles of length 120/60/30/15/1.
One recovers the $N^{\tau+1}=16^4$ possible initial
conditions when summing up all cycles together
with their respective multiplicities. The
most demanding system examined in this study 
is N16T5, for which
$16^6 = 2^{24} \approx 16.8\,\mathrm{M}$
initial conditions exists, together
with a corresponding number of distinct cycles.

\paragraph{Transformer details}
As a testbed, we use a decoder-only transformer
architecture with position-specific encoding
matrices, a variant denoted `cisformer'
\cite{gros2025small}. We work with
$N_{\rm con}$ context token and an
embedding dimension $d=N$, where $N$
is the basis of the NT task, see
Eq.~(\ref{NT_time}). A straightforward
orthogonal token embedding of the $N$ symbols
$\{0,1,2,\dots,N\!-\!1\}$ is implemented. 
Attention is causal. Expressive attention works
perfectly with standard positional embedding
schemes, which we did however not include. 
This simplifies the analysis of performance 
as a function of input length. 
A single transformer bilayer is used throughout 
this study, with one attention head per token.
Skip connections are present, with layer 
normalization being performed on entry, 
separately for the attention and the token-wise,
tanh feedforward layer. The width of the feed-forward
hidden layer is expanded by a factor four. 

\paragraph{Training}
Training is performed as a function of epochs, with 
each epoch consisting of $N_{\rm batch}$ predictions.
At the start of each epoch a new random NT series is
generated and encoded. The first $N_{\rm con}$ symbols
are then loaded into context space. For the underlying 
sequence we define a batch as the task to predict 
one-by-one the next $N_{\rm batch}=40$ symbols. 
A basic SGD optimizer is used during training, with 
momentum $\mu=0.8$ and a learning rate $\epsilon=0.02$.
Our aim is for a testbed that allows to study
relative performance, in particular as a function
of the complexity of the task, and not to attain
optimal performance by fine-tuning meta parameters.
If not stated otherwise, results shown are averaged 
over $N_r=16$ independent runs, 

\paragraph{Readout \& Testing}
The readout token is connected via a $dN_{\rm con}\times d$
matrix to the uppermost layer, with the task being 
to predict the next symbol of a NT time series. For the
loss function the basic squared difference is taken,
inference is greedy. During training, model performance 
is evaluated by asking the system to predict $N_g=50$
subsequent symbols of $N_{\rm test}=100$ distinct, 
randomly generated NT series. Once training is 
finished, larger numbers of $N_{\rm test}$ are 
used for the final evaluation, at least
$N_{\rm test}=10^4$.
The aim is to achieve 100\% accuracy.

\paragraph{Learning strategies}
For NT tasks, the symbols to be predicted 
are determined exclusively by two tokens situated
at fixed positions from the right end of the 
input sequence. As a matter of principle,
the complexity of the task would independent 
of $N$, $\tau$ and $N_{\rm con}$, the context 
length, if the models tested would
focus attention on exactly these two input
positions. Our results indicate that this
is not the case. Perfect performance is 
achieved nevertheless regularly, which implies 
that the respective strategies must 
be based on exploiting longer-range causal
correlations. When using this class of strategies,
task complexity does increase exponentially
both with the size of the basis and the delay,
$N$ and $\tau$, becoming easier on the other hand
when increasing context length.

\begin{figure}[t]
\centering
\includegraphics[width=0.98\linewidth]{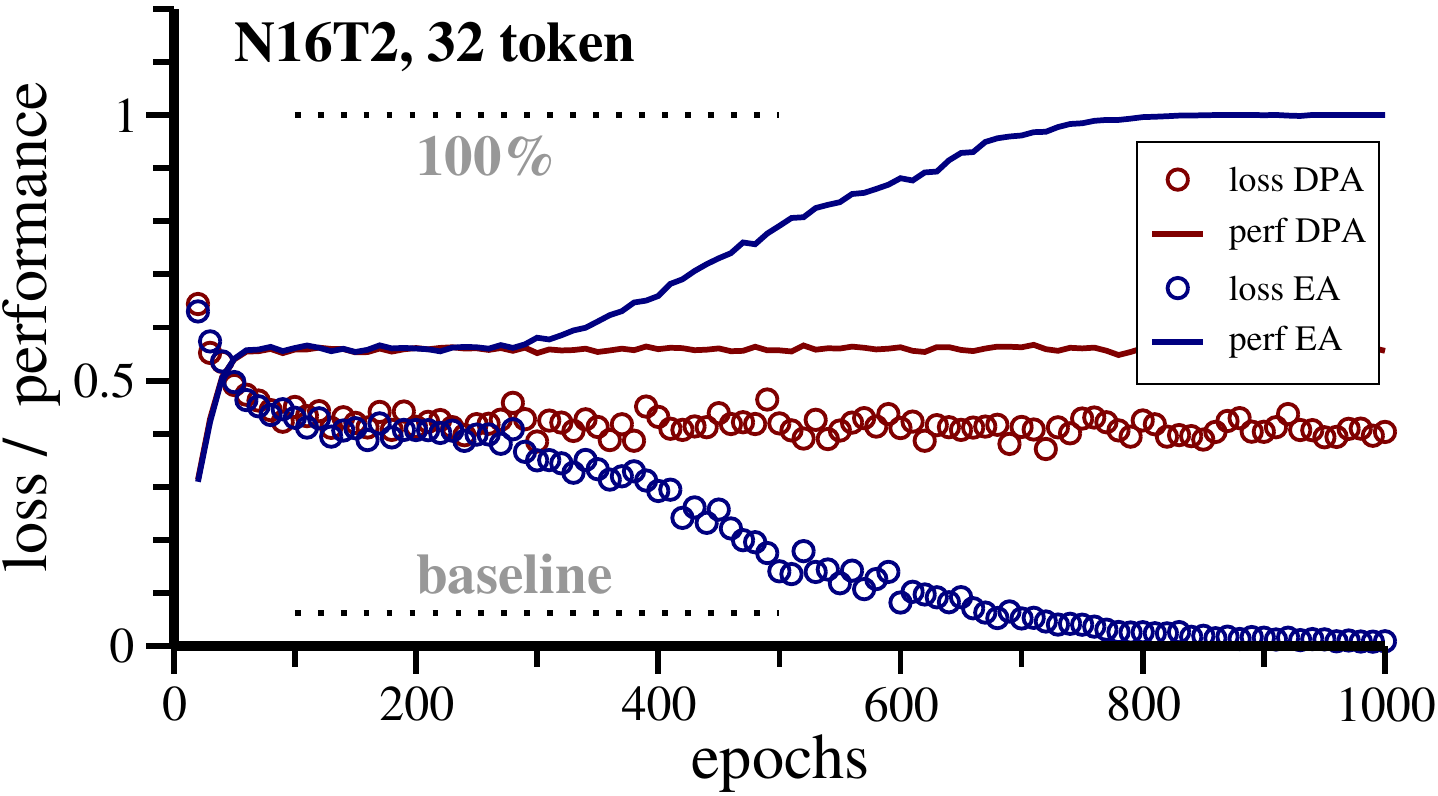}

\includegraphics[width=0.98\linewidth]{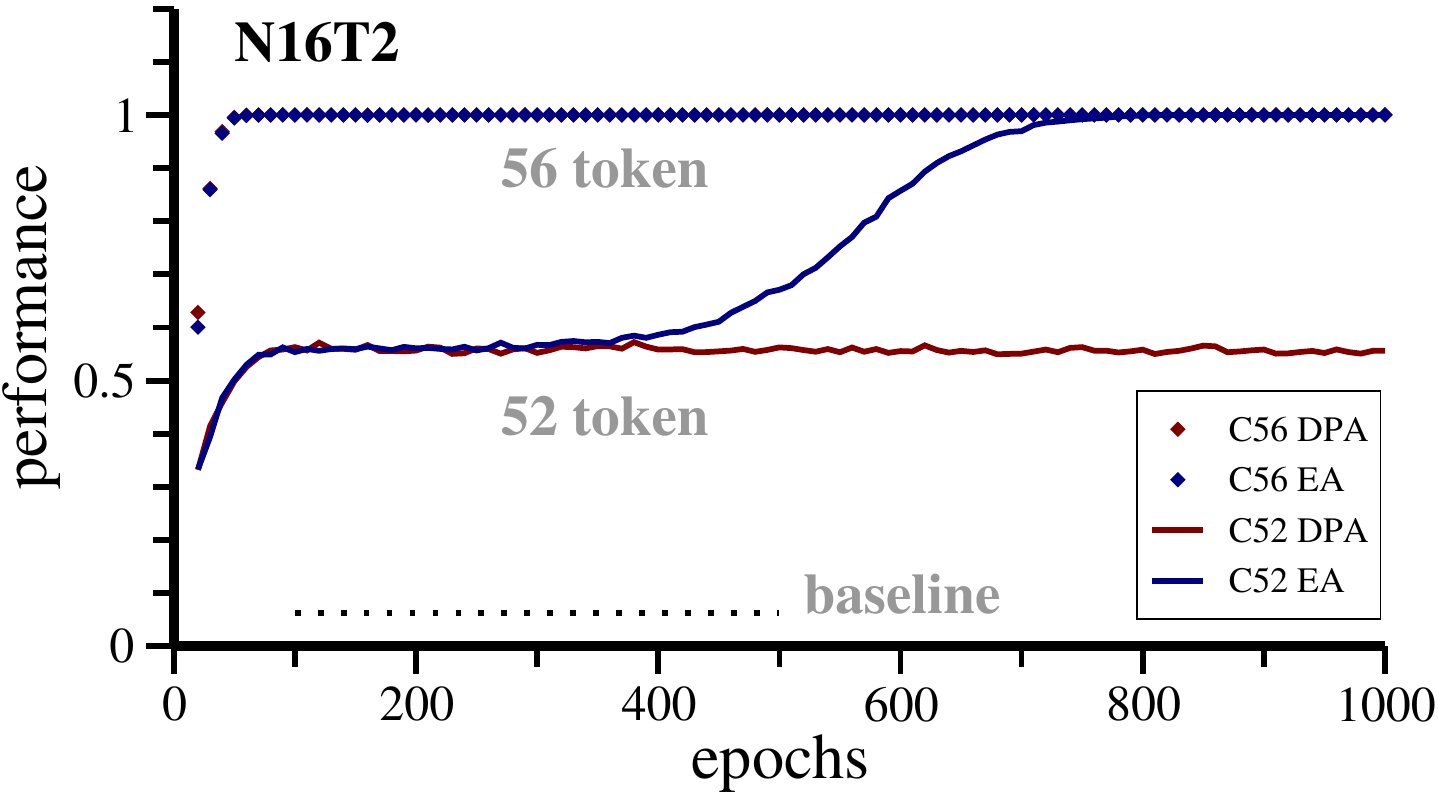}
\caption{For a NT task with basis $N=16$ and delay $\tau=2$,
the performance (lines) and the loss (rescaled by a
factor 0.75, circles). Result are for the standard dot-product
attention (DPA, maroon), and for expressive attention (EA, blue).
Performance is $1/16=0.0625$ (baseline) when predictions
are random, and $100\%$ when no errors are made. Shown
are results for a context length of 32 token (top panel) and
52/56 token (bottom panel), with DPA needing 56 token to achieve 
100\% performance. In this case, EA/DPA shown nearly
identical training behavior.
}
\label{fig:N16T2_C32}
\end{figure}

\section{Results}

A representative result is presented in
Fig.~\ref{fig:N16T2_C32}, where
simulations for $N=16$ and $\tau=2$ are
shown for both types of attention,
EA and  DPA. This is a task of moderate
complexity, with
\begin{equation}
64\cdot\underline{56} + 
16\cdot\underline{28} + 
4\cdot\underline{14} + 
1\cdot\underline{7} + 
1\cdot\underline{1} = 4096 = 16^3\,,
\label{N16T2_all}
\end{equation}
compare (\ref{N16T3_all}), which means that there are
64/16/4/1/1 periodic patterns of lengths
56/28/14/7/1. Three context lengths
are considered, $N_{\rm con}=32/52/56$.
\begin{itemize}
\item[--] 100\% performance\\
Eventually, both systems achieve 100\% accuracy,
predicting correctly 100 successive tokens for 
$10^4$ random starting N16T2 sequences. 

\item[--] structural traps\\
Lowering $N_{\rm con}$, eventually both
algorithms become trapped in a structural
local minimum. This is particularly evident
for $N_{\rm con}=32$, as shown in
Fig.~\ref{fig:N16T2_C32}. Both attention mechanism
rapidly improve their performance, reaching a 
plateau of about $55\%$, which is substantially
above baseline. At this point the DPA loss function 
stops improving, with progress slowing down
for EA. The phenomenology seen indicates that the 
enhanced expressivity of EA allows the system
to escape the local trap via a comparatively narrow
escape route.

\item[--] apparent emergence\\
As a function of system size, the DPA performance
rapidly increases from about $55\%$ to $100\%$. This
happens between $N_{\rm con}=52$ and $56$, but it 
would be wrong to interpret this performance jump 
as an `emergence phenomenon'. In fact, what happens is
that the structural local trap disappears eventually
when enlarging training space successively, an 
expected behavior.

\item[--] task complexity\\
For $\tau=2$, only $16^3=4096$ distinct
sequences exists for $N=16$. The task is
hence of modest complexity, given that
during training $N_{\rm batch}=40$ shifted
sequences are seen for any single epoch.
It is hence not surprising that the problem
can be solved within 100 epochs for both
models when the context length is large enough.
When reducing the number adjustable parameters,
both models require however larger training 
compute, when not failing completely. E.g.,
for $N_{\rm con}=16$ (not included in
Fig.~\ref{fig:N16T2_C32}), expressive attention
still converges to 100\% performance, needing 
however about 2000 epochs.
\end{itemize}
The results presented in Fig.~\ref{fig:N16T2_C32}
show in addition that training performance is
nearly identical for both attention mechanisms
when the problem at hand can be solved easily
with available resources. 

\begin{figure}[t]
\centering
\includegraphics[width=0.98\linewidth]{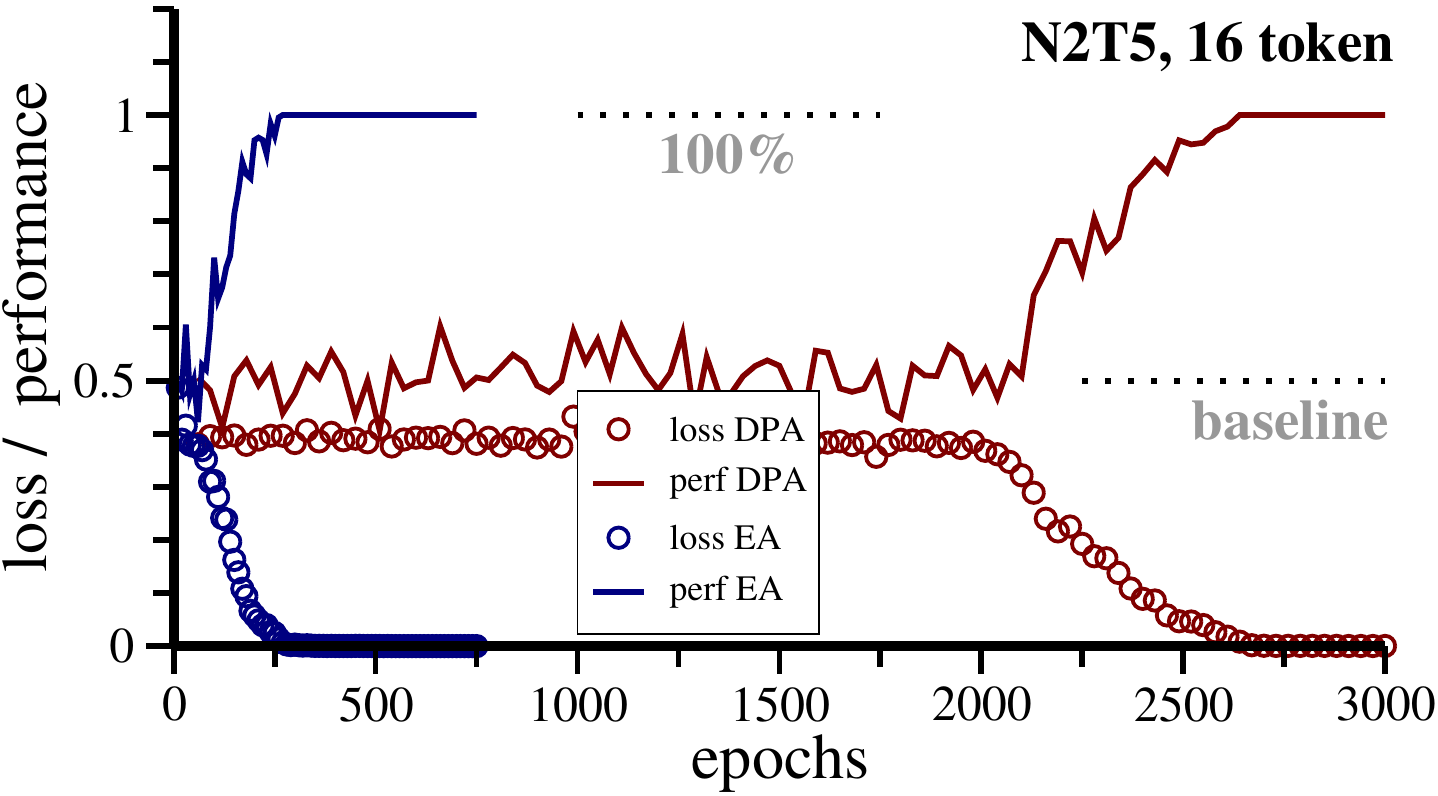}

\includegraphics[width=0.48\linewidth]{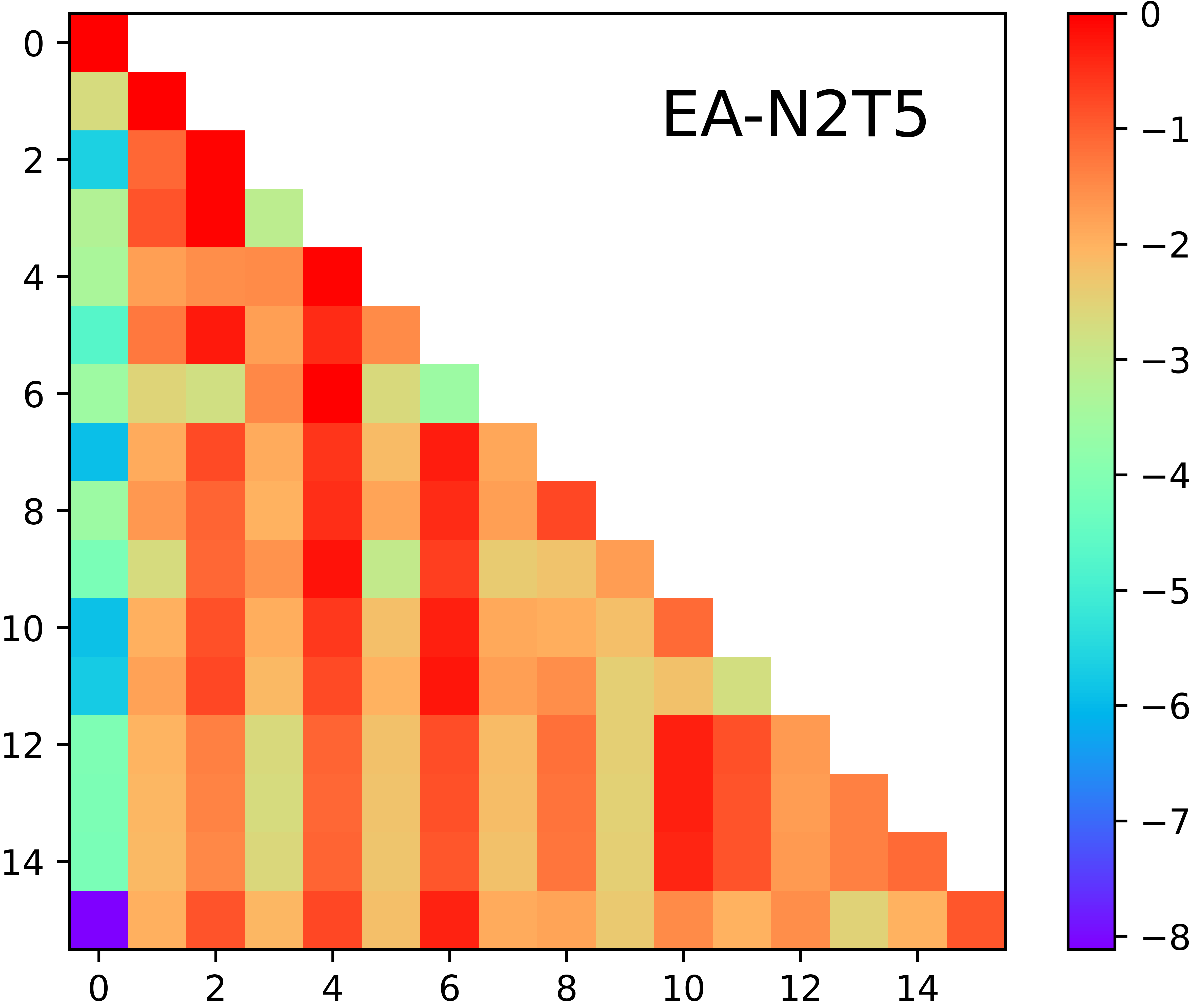}
\hfill
\includegraphics[width=0.48\linewidth]{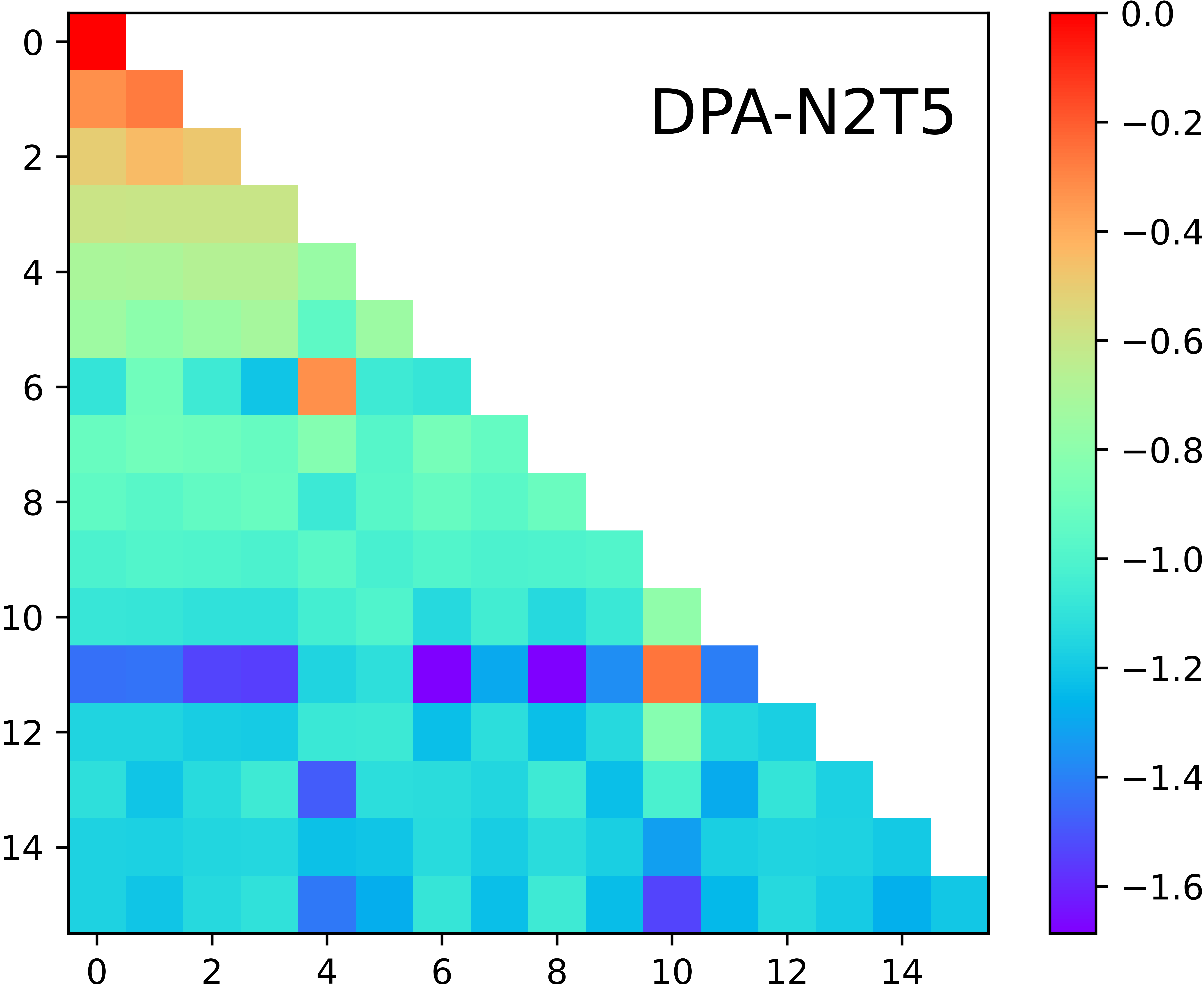}
 
\caption{Results for a NT task with a binary basis, 
$N=2$, and delay $\tau=5$. The model contains 802 
adjustable parameters for a context length 
$N_{\rm con}=16$, as used.
\textbf{Top:} Loss (rescaled by 0.75) and 
performance. For otherwise identical parameters,
dot-product attention (DPA) needs substantially 
longer to converge than expressive attention (EA).
\textbf{Bottom:} Snapshots of the respective
attention matrices. Typically,
EA (left) covers a larger dynamical range
(color-coding in $\log_{10}$ scale), than
DPA (right).
}
\label{fig:N2T5_C16_L1}
\end{figure}

\paragraph{Binary sequences}
It has been pointed out that an important 
aspect for our understanding of transformers 
is the study of their loss landscape 
\citep{hahn2024sensitive,choromanska2015loss}.
In this context we further investigate the
structural local minima observed in
Fig.~\ref{fig:N16T2_C32} by comparing the 
two attention mechanisms for the smallest 
possible embedding dimension, $d=N$, which 
is realized for the binary case $N=2$, 
compare (\ref{NT_time}). An additional 
question is here whether EA is advantageous 
even in this limit.  

We specialize to $\tau=5$, for which there are 
just two NT sequences, namely the default state 
`00000', and a cycle of length 63.
%
%
Together the $63+1=64=2^6$ possible initial conditions 
are covered, which makes N2T5 a seeming trivial 
task. For $N_{\rm con}=16$, and an embedding dimension 
$d=N=2$, overall model size is however modest, 
$N_{\rm model}=802$, which may induce pronounced local 
minima in the loss landscape. This is indeed the case, 
as the results presented in Fig.~\ref{fig:N2T5_C16_L1}
show. Averages over 16 runs have been taken.

For DPA one observes extended training spans during 
which performance fluctuates around the baseline, 
50\%, essentially without improving. For EA, the
equivalent training stage is substantially shorter.
The phenomenology observed indicates that learning
is dominated by a stochastic escape process
\citep{gros24complex}. Stochastic escape is present
when the additive effect of random fluctuating 
events allows systems to escape a local loss minimum.
However, further studies
would be needed to substantiate this hypothesis.
Of relevance for our studies here is the observations
that expressive attention may lead to a substantially
faster convergence than the classical dot-product mechanism.

Also included in Fig.~\ref{fig:N2T5_C16_L1} are
sample snapshots of the respective attention matrices.
Visual inspection indicates that both models make
use of the entire input sequence, viz that attention
is not focused predominantly on the two causal
token, as defined by (\ref{NT_time}).

\begin{figure}[t]
\centering
\includegraphics[width=0.98\linewidth]{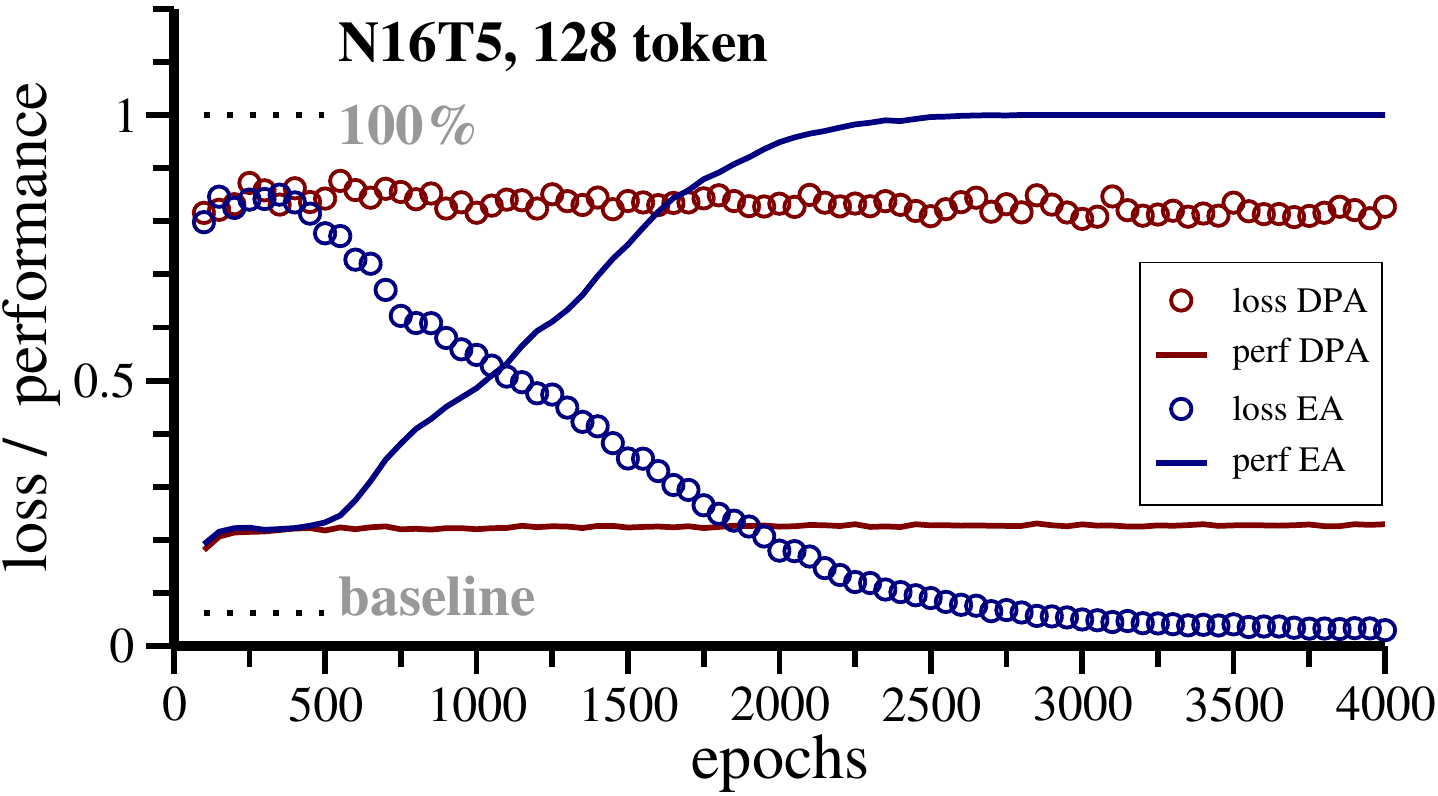}

\caption{For a NT task with basis $N=16$, delay $\tau=5$,
and 128 token, the performance (lines) and the loss 
(rescaled by a factor 0.75, circles). Qualitatively, 
the results mirror the ones presented in
Fig.~\ref{fig:N16T2_C32}, both for 
the standard dot-product attention (DPA, maroon), 
and for expressive attention (EA, blue). 
When training for 2000 epochs, models see about
0.5\% of the $16.8\cdot10^6$
possible sequences.
}
\label{fig:N16T5_C128}
\end{figure}

\paragraph{Large sequence spaces}
The number $N_{\rm seq}=N^{\tau+1}$ of 
distinct NT sequences has been modest for 
the results presented in
Figs.~\ref{fig:N16T2_C32} and \ref{fig:N2T5_C16_L1}. 
As a consequence, exact solutions in the form 
of learned one-by-one mappings $N_{\rm seq}\to N$ 
are conceivable, at least as a matter of principle.
This is however not any more the case for large
$N_{\rm seq}$.

In Fig.~\ref{fig:N16T5_C128} we present results
for N16T5. Model size is about $4\cdot10^5$
for $N_{\rm con}=128$, in 
terms of adjustable parameters, 
which is substantially smaller
than the number of distinct NT sequences,
$N_{\rm seq}=16^6 \approx 16.8\cdot10^6$.
During training, $N_{\rm train}=0.8\cdot10^5$ 
training sequences are presented within the 
first 2000 epochs, which is about 0.5\% of $N_{\rm seq}$.
It is hence likely that the generative mechanism
has been encoded in one form or another once 
performance is either perfect, or near to 100\%.
As before,  we used $5\cdot10^5$ predictions
for the evaluation.

On a qualitative level, the results shown in
Figs.~\ref{fig:N16T5_C128} and
\ref{fig:N16T2_C32} are equivalent.
Both attention mechanisms converge rapidly
to a heuristic strategy with a non-trivial
performance, respectively of the order of
0.55 for N16T2 and 0.2 for N16T5. In both
cases, DPA remains stuck in the corresponding
local minimum, with EA escaping at a reduced
pace. 


We did not determine at which context length
DPA would converge. For $N_{\rm con}=256$, DPA 
remains in a local trap, but with an improved 
performance, of roughly 0.44. Increasing the context 
length further, to $N_{\rm con}=512$, we find 
that the training of both DPA and EA becomes 
unstable when our standard learning parameters are 
used, namely $\mu=0.8$ and $\epsilon=0.02$. 
One could stabilize training by adjusting momentum
and learning rate, which is however outside the 
scope of our comparative analysis.

\paragraph{Mixture of tasks}
The results presented till now concerned
setups containing only a specific single 
tasks. We performed simulations also
for mixture of tasks, finding that
expressive attention outperforms classical
dot-product attention generically also in this
case. We define a new task variant, NT-S,
\begin{equation}
\hbox{\sc NT-S:}\quad
x(t) = \left(\sum_{\Delta T=1}^{\tau+1}x(t-\Delta t)\right)\%N\,,
\label{NT-S_time}
\end{equation}
which corresponds to summing up the $\tau+1$ preceding
token, modulo $N$. The two versions, NT-S and NT are 
identical for $\tau=1$, compare (\ref{NT_time})
and Fig.~\ref{fig:NT_illustration}, but not 
for $\tau>1$.

\begin{figure}[t]
\centering
\includegraphics[width=0.98\linewidth]{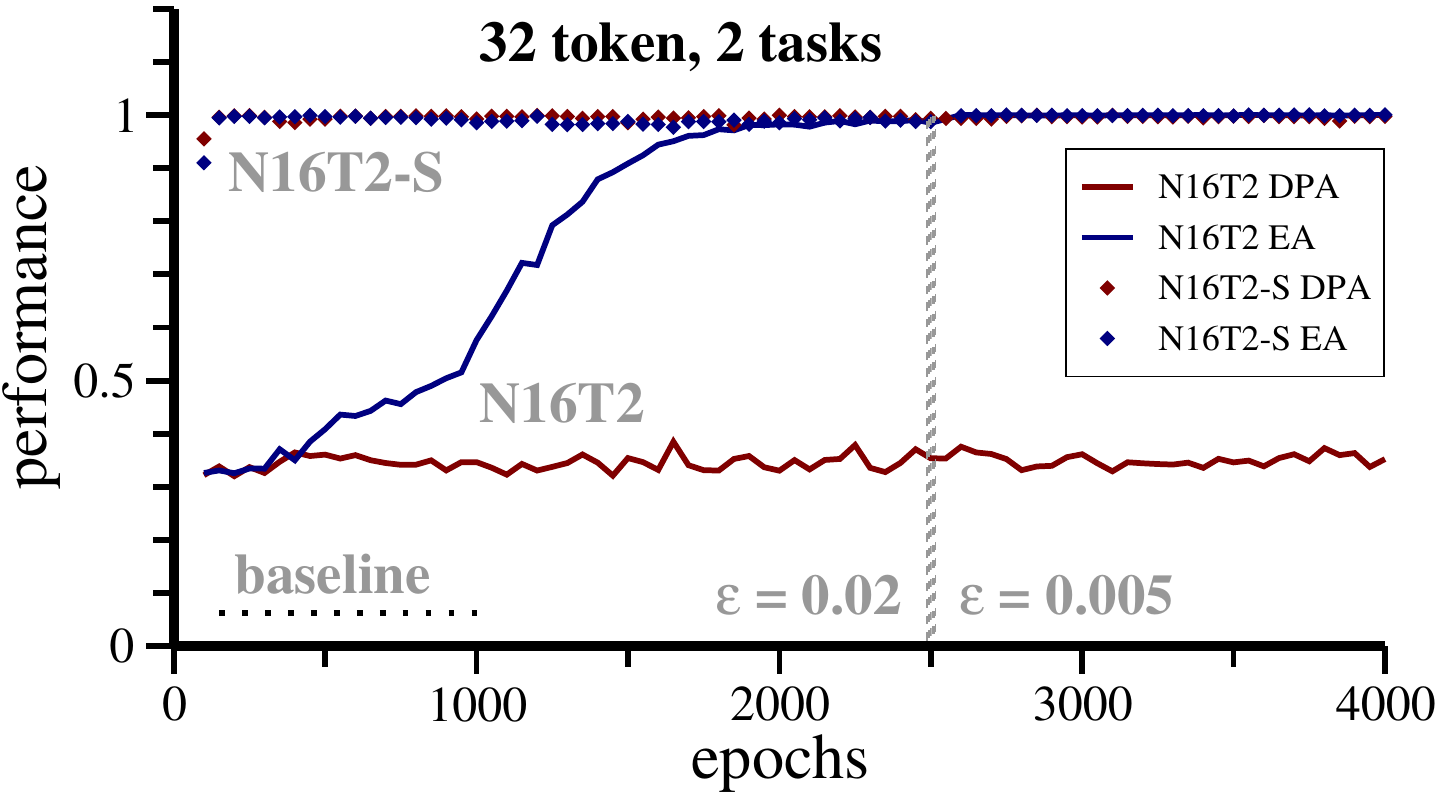}
\caption{For a mixture of two tasks, N16T2 and N16T2-S,
the performance of EA (blue) and DPA (maroon). Compare
Fig.~\ref{fig:N16T2_C32} for the case when only N16T2 
is trained. After 2500 epochs (shaded vertical line),
learning speed is reduced by a factor four.
}
\label{fig:N16T2_C32_dual}
\end{figure}

Results for $N_{\rm con}=32$ are given in
Fig.~\ref{fig:N16T2_C32_dual}. During training
a 50/50 mixture of N16T2 and N16T2-S autoregressive
prediction tasks are presented as such, viz without
further information. For a given prompt, which could
be either a N16T2 or N16T2-S sequence, the system 
needs to determine on its own both the delay $\tau$ 
and the type of the task at hand.

NT-S tasks typically lead to shorter cyclic patterns
than the corresponding NT task, which makes them easier
to learn. For example, the mean cycle length is 47.6 for
N16T2, but only 23.8 for N16T2-S. Both systems, EA and DPA 
have consequently no problem to achieve 100\% accuracy on
N16T2-S as a single tasks. When combined with N16T2,
both systems achieve still above 99\% accuracy, as
shown in Fig.~\ref{fig:N16T2_C32_dual}. Also of
interest is the reduction of the steady-state N16T2 
performance in the mixed-task scenario, from about 
0.55 to 0.34, which is also the value achieved by 
DPA. 

In our setup, expressive attention is able to
solve both tasks to about 99\%, Given that N16T2 
is shown only 50\% of the time, it is not 
surprising that learning is prolonged, as
evident from Fig.~\ref{fig:N16T2_C32_dual}. 
As an experiment we lowered learning speed after 
2500 epochs, finding that N16T2 performance 
increases a bit, by about 0.5\%. Interference 
between competing tasks is reduced at lower 
learning rates.

Tasks are acquired consecutively in order of
difficulty, as shown in Fig.~\ref{fig:N16T2_C32_dual}, 
which can be interpreted as an instances of unsupervised
curriculum learning \citep{bengio2009curriculum}. 
Our results can be seen also as an explicit example of
successive learning of task quanta, as postulated in
the quantization theory of neural scaling
\citep{michaud2024quantization,neumann2022scaling,neumann2024alphazero}.

\paragraph{Rare events}
As a third variant, we introduce via
\begin{equation}
\hbox{\sc NT-R:}
\quad\quad
\left\{\begin{array}{ll}
\mathrm{NT\!-\!S} & \mathrm{if} \ \ \ x(t-\tau-1)=0 \\[0.5ex]
\mathrm{NT}   & \mathrm{otherwise}
\end{array}\right.
\label{NT-R_time}
\end{equation}
a logical if-statement. The generating algorithm
remains deterministic, switching from NT
to NT-S when a given condition is fulfilled,
here when $x(t\!-\!\tau\!-\!1)=0$. For an observer
this corresponds to a rare event that occurs with 
a probability of 1/16=0.0625 when N=16.

\begin{figure}[t]
\centering
\includegraphics[width=0.98\linewidth]{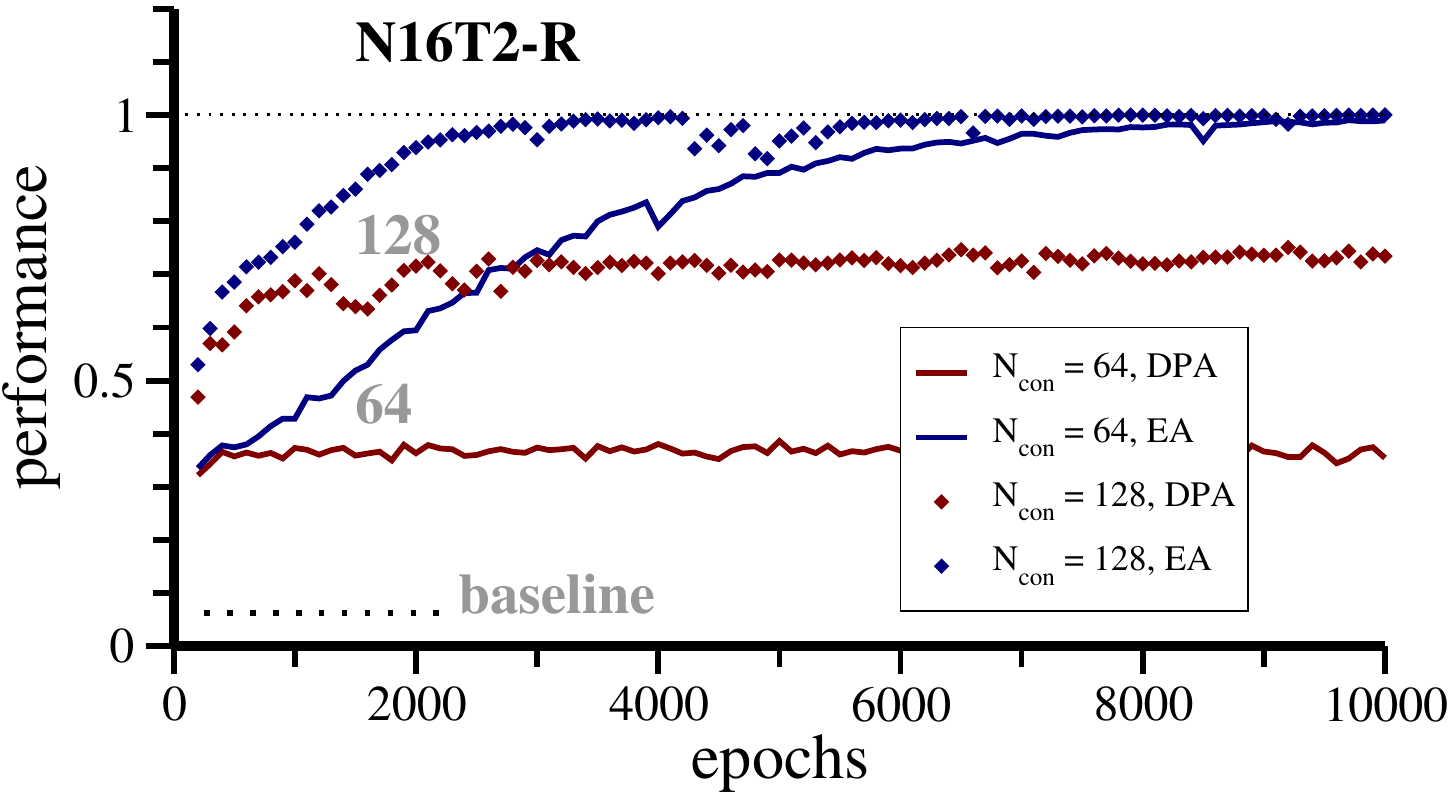}
\caption{A N16T2 task switching conditionally
to N16T2-S, as defined by (\ref{NT-R_time}).
Data as averaged over 16 runs.
}
\label{fig:N16T2_C64_128_rare}
\end{figure}

Determining the presence of an isolated hidden 
logical statement, as defined by (\ref{NT-R_time}), 
can be an exceedingly demanding task. In 
Fig.~\ref{fig:N16T2_C64_128_rare} results for
$N_{\rm con}=64/128$ are presented. For
these two context lengths N16T2 is 
comparatively easy, as shown in 
Fig.~\ref{fig:N16T2_C32}, which implies
that a certain level of performance should
be attainable in any case. For DPA this level
about 0.36 for $N_{\rm con}=64$, doubling
to 0.72 when the context length is raised
to 128.

The data presented in Fig.~\ref{fig:N16T2_C64_128_rare} 
demonstrate that it is difficult also for 
expressive attention to isolate a lone 
if-statement, as defined by (\ref{NT-R_time}).
Training progress is slow, achieving in the
end however an average accuracy of 98.8/99.9\%,
respectively for $N_{\rm con}=64/128$. As
usual, averages over 16 random initial conditions 
have been are taken. The majority of runs achieves 
however perfect performance. We did not investigate
the cause of the non-monotonic events showing up
during training, which may be due to periods
with an increased clustering of rare events.

\section{Discussion}

Obtaining encouraging results, we presented 
a first evaluation of expressive attention (EA),
focusing on single-layer models with a single
attention head. We also investigate models 
with two or more layers and/or attention heads, 
together with different learning speeds and/or 
momentum. All results are found to be robust. 
Additional investigation would however necessary
for a full assessment of the potential of EA,
e.g.\ in the realm of natural language processing 
\citep{kaddour2023challenges,yang2024harnessing}.
or within the context of formal languages
\citep{strobl2024formal,abbe2023sgd}.
Generally, we expect that expressive attention 
will fare at least as well as dot-product 
attention (DPA), a presumption that is based
on the respective design principles. For DPA,
large and small attention weights are constrained 
to a one-dimensional manifold within the space 
of attention heads, which is not the case for EA.
The size of the performance boost obtainable
when substituting DPA by EA may depend
however strongly on the use case.

\paragraph{Heuristics}
Repeatedly we observed that initial performance 
gains flatten out rapidly. This happens in particular
for small model sizes, but also for DPA in cases
when the available resources are sufficient for 
EA to solve the problem at hand exactly. We argued
that the resulting stationary performance plateau 
indicates the presence of a local minimum in 
loss landscape. From this trap models may escape
either by an embedding into into a larger dimensional
parameter space, or by an improved design. For all
cases the level of the observed stationary performance 
plateau was independent of initial conditions and 
training details. This led us to the conclusion that the
associated local minimum in loss landscape is structural, 
viz that it corresponds to a heuristic strategy. 
To examine how this heuristic strategy works would
an interesting research question, which is however
beyond the scope of the present study.

\paragraph{Reasoning}
Inductive reasoning is one of the big challenges 
of large language models \cite{li2024llms,su2023llms}. 
It is encouraging that one of the building blocks, 
the logical if-statement included, can be extracted 
and encoded when effective attention is used. Our
analysis, which involved logical task switching
within the NT-R framework (\ref{NT-R_time}),
is however only a first indication and it
remains to be seen to which extent EA may raise
performance in this field.

\subsubsection*{Acknowledgments}
We thank Michael Hahn for discussions.

\end{document}